# The influence of Chunking on Dependency Crossing and Distance

QIAN LU[1], CHUNSHAN XU[1,2] and HAITAO LIU[1, 3(a)]

[1] *Department of Linguistics, Zhejiang University - CHN-310058 Hangzhou, China*
[2] *School of Foreign Studies, Anhui Jianzhu University - CHN-230601 Hefei, China*
[3] *Ningbo Institute of Technology, Zhejiang University, CHN-Ningbo, 315100, China*

PACS 89. 75. Hc – Networks and genealogical trees

PACS 05. 40. Fb – Random walks and Levy flights

PACS 89. 90. +n – Other topics in areas of applied and interdisciplinary physics

**Abstract** – This paper hypothesizes that chunking plays important role in reducing dependency distance and dependency crossings. Computer simulations, when compared with natural languages，show that chunking reduces mean dependency distance (MDD) of a linear sequence of nodes (constrained by continuity or projectivity) to that of natural languages. More interestingly, chunking alone brings about less dependency crossings as well, though having failed to reduce them, to such rarity as found in human languages. These results suggest that chunking may play a vital role in the minimization of dependency distance, and a somewhat contributing role in the rarity of dependency crossing. In addition, the results point to a possibility that the rarity of dependency crossings is not a mere side-effect of minimization of dependency distance, but a linguistic phenomenon with its own motivations.

**Introduction.** – Language used in communication is invariably presented linearly, one unit after another, which is regarded as one of its fundamental property [1]. However, there is always a sytactic tree structure underlying a one-dimensional linear sentence, a structure underpinning both the production and the comprehension of this sentence [2,3]. Therefore, language processing consists, to a considerable degree, in the transformation between the syntactic tree structure and the one-dimensional linear arrangement. What properties can be found in the tree structure of language? What mechanisms constrain the transformation of tree structure into linear structure? The answers to these questions, which may well require researches based on statistical physics and computer simulation, probably will shed much light on how human language operates.

In terms of dependency grammar, the structure of a sentence can be visualized as a hierarchical dependency tree, whose nodes (vertices) are words, linked to one another by directed edges (dependency relations) [2,3]. Such a hierarchical tree must be ultimately arranged into a linear sequence, for the purpose of spoken and written communication. So far, researches have repeatedly observed two phenomena in the linear realization of hierarchical dependency structure: the minimization of dependency distance (the number of intervening words) between two syntactically related words [4-13], and the rarity of crossing dependency relations [14,15]. Liu [5] has compared dependency distance of 20 natural languages with that of two different random languages, and pointed out that dependency distance minimization seems to be universal in human languages. Ferrer-i-Cancho has theoretically analyzed these [8,9]. A recent study based on 37 languages has obtained similar findings[11]. Since dependency distance is held as cognitively related to language processing load [16], the minimization of dependency distance is probably a result of the principle of least effort [17]. In addition, it is argued that that the rarity of crossing dependencies is simply a by-product of the pressure to minimize dependency distance and cognitive cost in language processing, having little to do with the syntax of the language [7-10]. Similarly, some studies find that dependency distance will significantly increase if dependency crossings are permitted, and suggests that reducing dependency crossings is probably an important means to restrain dependency distance [4,5].

Dependency distance and crossings are closely related, and in human languages both seem to be subject to minimization. Ferrer-i-Cancho [9,10] has theoretically proven that, for sufficiently short dependency lengths, the probability that two edges cross decreases as their length decreases. However, Liu has found that projective random language (i.e. without any crossing dependency) has significantly longer mean dependency distance than natural langauge [4,5]. Therefore, the short MDD of natural languaes can not be wholly ascribed to the rarity of crossing dependency : there must be some other mechanisms contributive to it.

Previous researches on this issue have mainly focused on the linear distance between syntactically related nodes, or the crossing dependencies, neglecting, somewhat, the role of the hierarchical syntactic structure which, we believe, probably have much influence on the linear ordering of words in a sentence. Such a hierarchical structure implies another basic operation in language processing, namely, chunking. In fact, one defining characteristic of human languages is duality [18], that is, smaller units at lower level combine to form bigger units at higher level. In other words, words may combine with other words to form chunks, which may in turn combine with other chunks to form even bigger chunks until a sentence is established. In terms of dependency grammar, some daughter words may depend on one father word, forming a chunk, which behaves as a whole and syntactically relates to another chunk, via the dependency between the governing words of both chunks, to form even bigger chunks. Since a dependency tree is hierarchical, chunking usually operates at many levels in a bottom-up fashion, from words to clauses. In short, it may be assumed that, in many cases, the hierarchical structure bears on how a string of words are grouped into chunks.

Chunking therefore may serve as an interface between hierarchical dependency trees and the linearization of these trees. In fact, chunking has been found playing significant roles in human information processing [19], widely used in the construction of automatic syntactic parsing systems [20]. Then a question follows: does chunking, which bears on hierarchical syntactic structure, contribute to the minimization of dependency distance and crossings in human languages? To answer this question, this paper simulates, with multilayer random walk algorithm, the linear chunking of dependency trees, and conducts comparison with natural languages, focusing on the following issues: does chunking in linear sequence, contribute to the minimization of dependency distance and the rarity of crossing dependencies. The second section is devoted to the theory and the method used in our study. The third section discusses the potential effect of chunking on dependency distance and crossing dependencies. The fourth section is the conclusion.

**Method.** – Dependency grammar holds words as the fundamental syntactic units, linked via dependency relations into complete syntactic constructions [2, 3]. The structure of a linear sentence can be visualized as a dependency tree, which, according to the basic principle of dependency grammar, should be a connected tree with one single root node [21]. In Fig. 1(a), the sentence "I like red apple" has a root node: "like"; the directed arcs above this sentence indicate the dependency relations among words, and parts of speech are marked underneath every word.

These basic properties of dependency trees have been formally described by Liu and Hu [22]. Rarely, the principle of continuity might be violated, that is, one dependency relation crosses another, which is labeled as type-I crossing, as illustrated in Fig. 1(b); or, one dependency crosses the root node, which is labeled as type-II crossing, as illustrated in Fig. 1(b). Crossing has much to do with word order: the sentence in Fig. 1(b) is the same as that in Fig. 1(a) except for change in the postion of "red", and thus becomes ungrammatical.

Hudson [16] first defined the dependency distance and associates it with the cost of language processing. [4,8] produced similar formula to calculate the mean dependency distance (MDD), as a metric of language processing difficulty. Our research adopts their approach, and focuses on the impact of chunking on the MDD.

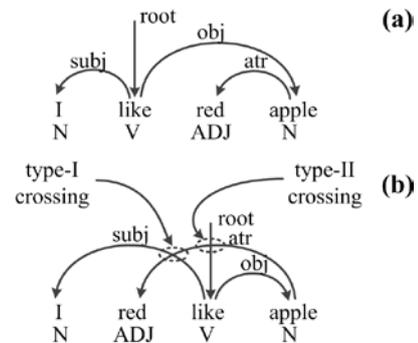

Fig. 1: (a) A well-formed dependency tree. (b) A dependency tree with crossing arcs.

In this paper, the dependency structure of sentence is a tree[2,3]; chunk is defined as the segment composed of a governor and its dependents, and, if any, these dependents' dependents. To be precise, a chunk would be labeled with the root of a subtree and contain all the nodes from that root to the leaves, including at least one node. Mathematically, chunks are a partition of the set of nodes (words) that constitute the sentence, furthmore, two chunks cannot contain the same nodes. At the same time, situation of chunks be embedded in each others is excluded. The MDD of a sentence with *n* words can be calculated with the following formula:

$$MDD_S = \frac{1}{n-1}\sum_{i=1}^{k}(|cdd_i|+|ldd_i|) \qquad (1)$$

In this formula, $k$ refers to the number of chunks in a sentence; $cdd_i$ refers to the sum of the internal dependency distances within the $i$th chunk; $ldd_i$ refers to the dependency distance between chunks, which is defined as the distance between the root of the i-th chunk and the governor of the root of the i-th chunk.

We define a chunk as a sub-tree, a part of the dependency tree of a sentence. The expected MDD of a linguistic construction is in proportion to the length of the construction [8,9]. Therefore, it can be inferred that the MDD of chunks is smaller than that of the entire sentence. At the same time, chunking inevitably leads inter-chunking dependencies, which are likely to be of long distance. According to formula (1), the MDD of a sentence is most susceptible to the dependency distances among chunks. Therefore, MDD may also bear on the length of chunks.

To clarify the impact of linear chunking with different lengths on dependency distance and crossings, we generate random trees by algorithm 1 that segments linearly ordered nodes into chunks of certain lengths, then dependencies within and among chunks are built with algorithm 2.

Algorithm 1:
1) For a linear sequence: $S = 1, 2, \cdots, n$, with a set of nodes: $V_c = \{1, 2, \cdots, n\}$; a dependency tree is randomly generated with the following initial state of tree: the set of tree nodes $G_c = \{\ \}$, and the set of tree edges $E_c = \{\ \}$;
2) With the maximum length of chunk set as MAX(<=sentence length), and the minimum as MIN(>=1), the linear sequence of nodes $S$ is divided into $k$ chunks: $C_1, \cdots, C_k$;
3) For each chunk, a dependency sub-tree is randomly generated with algorithm 2: $G_1, \cdots, G_k$; the root nodes of these sub-trees are the heads of each chunks: $Head_1, \cdots, Head_k$, of which one is randomly chosen as the root of the entire linear sequence; then the nodes and edges of sub-trees $G_1, \cdots, G_k$ are added to $G_c$ and $E_c$;
4) Among chunks are generated $k$-$1$ edges $\langle C_i, C_j \rangle$, $i \neq j \in [1,k]$, with $C_i$ governing $C_j$ ; in $C_i$, a node $m$ is randomly chosen as the governing node that connects to the head node of $C_j$; then edge $\langle m, Head_j \rangle$ is added to $E_c$;
5) When $k$-$1$ edges are generated between chunks, output $G_c$ and $E_c$; the algorithm ends.

Random directed tree generating algorithm (algorithm 2) [22,23] is used in step 3 and 4 of algorithm 1 to generate dependency tree satisfying the requirements of single-governor, single-root and connectedness.

The sizes of chunks range randomly between [MIN, MAX]. This interval can be set according to our needs so as to explore the possible relations between chunk size and dependency distance.

Combined with projective (continuous) random tree generating algorithm[22], above approaches can produce four types of random trees:
1) A single root random acyclic connected tree whose nodes has only one parent apeice. This random tree is generated by algorithm 2, allowing crossing dependencies (e.g., Fig. 2(a)).
2) A single root random acyclic connected continuous tree whose nodes has one parent apeice. This random tree is generated by algorithm 3, and prohibits crossing dependencies, as shown in Fig. 2(b).
3) A single root random acyclic connected tree with chunks, whose nodes have only one parent apeice. This tree is generated by algorithm 1. There is no continuity constraint in this algorithm, and thus crossing dependencies may occur (e.g., Fig. 2(c)).
4) A single root random acyclic connected continuous tree with chunks, whose nodes have only one parent apeice. This tree is generated by algorithm 4, that is, by replace algorithm 2, which is contained in algorithm 1, with algorithm 3, so as to put constraint of continuity on tree generation and prevent crossing dependencies (e.g., Fig. 2(d)).

Four random treebanks are established, namely RL1, RL2, RL3, RL4, each comrpising exclusively only one type of above randoms trees. Sentence length in each treebank ranges between 2-100, and for each sentenc length, 5000 random trees are generated. Instead of using rules extracted from real language to generate dependency trees [12,13], we randomly generate dependency trees to exclude the influence of syntactic rules for the purpose of an objective probe into the effect of chunking on dependency distance and dependency crossings.

In addition, we also statistically investigate a treebank of Mandarin Chinese [24], which serves as a baseline for comparison. The texts of this treebank come from People's Dailly, containing 14463 sentences and 336138 words.

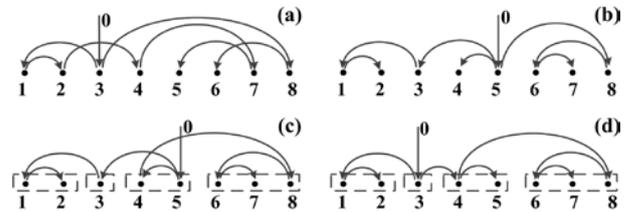

Fig. 2: (a) A random tree with crossing arcs. (b) A random tree with no-crossing arcs. (c) A random tree with crossing arcs based on linear chunking. (d) A random tree with no-crossing arcs based on linear chunking.

**Results and discussion.** – The relations between MDD and sentence length (SL) is shown in Fig. 3. The reference line on X axis indicates the mean sentence length of natural language (NL), which is 23. The reference line on Y axis indicates the MDD of NL, which is 3.79.

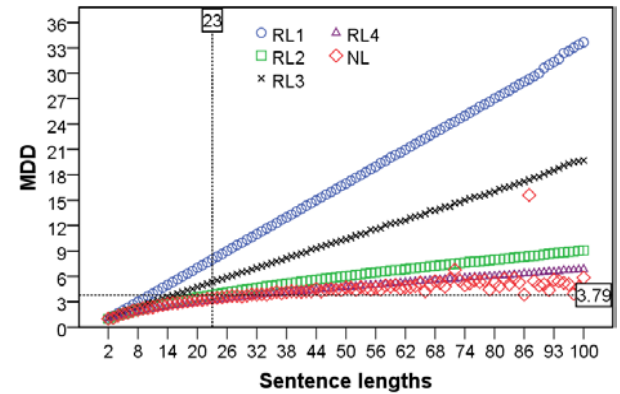

Fig. 3: Relations between MDDs and sentence lengths.

Fig.3 reveals that the MDD of RL3 is longer than the MDD of RL2, which suggests that, compared with continuity,

chunking has weaker ability to reduce MDD. Despite that, it is clear that both continuity and chunking have effect on MDDs of random languages (RL), because the MDDs of RL2 and RL3 are both shroter than that of RL1. Therefore, depenency relations between chunks (*lddi* in formula (1)) do not increase MDD of a tree in general.

Statistical studies of various languages have repeatedly pointed to rarity of crossing dependencies in natural language[5,7]. In the Mandarin Chinese treebank of our study, crossing dependencies are also absent. These findings suggest that the rarity of crossing dependencies in natural languages may be due to the pressure for short dependency distance. Fig. 3 also shows that the MDD of continous random language (RL2) is very close to that of natural language. However, when sentence length is over 20, the difference between MDDs of RL2 and NL steadily widens. It can then be inferred that dependency distance minimization (DDM) cannot be exclusively attributed to the rarity of crossing dependencies: there might be other mechanism that contribute to DDM, especially in the cases of long sentences and chunking is probably one of them. As shown in Fig.3, when SL is over 20, the MDD of RL4, which is both continuous and chunked, is almost identical with that of NL. Chunking is closely related to the duality of languages, that is, the process of iteratively combining smaller units into larger ones, which is held as one fundamental feature of human languages. In short, chunking serial sequence is probably another means to reduce dependency distance.

In addition, statistical tests indicate a significant correlation between sentence length and MDD (Pearson Correlation = 0.94, P-Value = 0). That is, the longer a sentence is, the longer MDD it will have, which provides another evidence for the purported relation between sentence length and MDD [6,9]. Therefore, long sentences, especially those longer than 100, will cause severe comprehension trouble. Hence, as a self-adapting system, language will evolve certain mechanisms to cope with this problem. Chunking these long sentences into short sentences might be one of them, which probably marks the interface between syntax and discourse.

Above discussion points out that chunking can play a significant role in reducing MDD. However, another question arises: can chunking reduce MDD to that of NL, without the constraint of continuity?

We investigated the possible relation between chunk size and MDD by controlling chunk size in two way. Fig.4(a) indicates the relation between MDD and chunk size when maximal chunk size is controlled. That is, chunk size may vary within a certian range [1,*n*], and the maximum, i.e., *n*, is controlled so that we can observe how MDD is affected by the maximal size of chunks. Fig. 4(b) indicates the relation between MDD and chunk size when chunk size is controlled. That is, there is no variation of chunk size; all chunks are of the same length. In Fig. 4, We choose several sentence lengths based on power of 2. Thus, the sentence lengths would be 2, 4, 8, 16, 32, 64. We use SL2, SL4, SL8, SL16, SL32, SL64 to represent these lengths. As can be seen in Fig. 4, when chunk size is increased gradually for 1 to 64, the MDDs at different sentence lengths all change accordingly. Fig.4 indicates that there is a minimum of MDD that can be reached by increasing chunk size from 1 to 64. Before reaching the minimum, increasing chunk size reduces MDD, but after reaching it, increasing chunk increses the MDD.

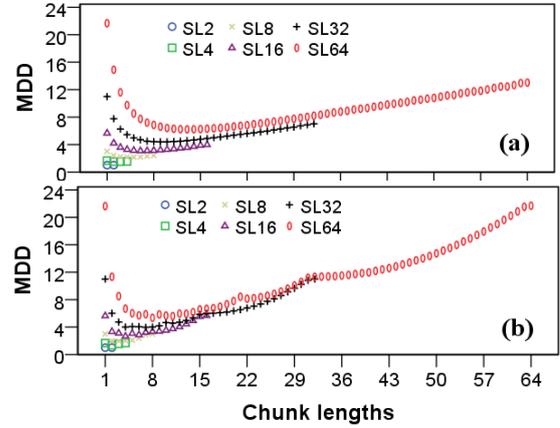

Fig. 4: Relations between MDDs and chunk lengths. (a) Random chunk lengths. (b) Fixed chunk lengths.

Fig.4(a) reveals that when sentence length is 2, 4, 8, 16, 32, 64, the minimum MDD can be reached if maximal chunk size (n) is set respectively as 1, 3, 5, 6, 10, 14. Fig.4(b) shows that for these sentence lengths, the minimum MDD can be reached if chunk size is set as 1, 2, 3, 4, 7, 8. These findings suggest that more chunks do not necessarily mean less MDD. Linguistically, that means more hierarchical levels in a synactic tree will not necessarily reduce comprehension difficulty (MDD).

We obtained the minimal MDDs of random sentences that are chunked, and compared it with continous random language(RL2) and natural language (NL), as illustraded in Fig. 5. In this figure, RL3_RC and RL3_FC indicates the minimal MDDs of random sentences when maximal chunk size is controlled and when all chunks are of the same length, as we formerly mentioned in Fig 4 (a) and Fig 4 (b).

Obviously, proper chunking can significantly reduce the MDD of a linear sequence. Nodes, can only govern or depend on other nodes in the same chunk, which puts a limit on the number of long distance dependencies, since only one dependency relation is permitted between two chunks. Normally, for sentences whose length is 23, long dependency distance is much more likely to be found between chunks than within them, when chunks are composed of 4-7 nodes. Proper chunking means many short intra-chunk dependencies and several potentially long inter-chunk dependencies, making for shorter MDD than an un-chunked sequence, where long distance dependencies are not limited by chunks.

If the entire sequence can be seen as equivalent to a sentence, the chunks may be held as somewhat similar to clauses. Interestingly, existent studies [25] have reported average clause lengths as ranging between 4 and 8, rather close to the chunk lengths of 4-7, which are found in our study as most potent in limiting the MDD (16<SL<32). However, this study [25] reported that there are averagely two clauses in a sentence, while our study revealed that there are optimally 1-8 chunks in a sequence (SL≤64). Why are there less and slightly longer clauses in a sentence of real natural languages? This question is probably worthy of further investigation.

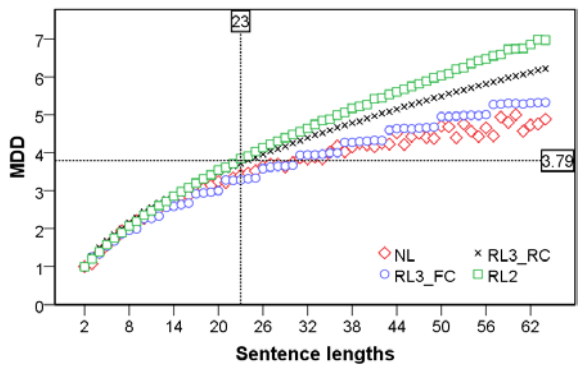

Fig. 5: Relations between sentence lengths and MDDs.

The data have shown that chunking alone may limit MDD to that of natural languages even without the interference of syntactic rules. But another question still wants answer: does chunking also reduce the number of crossing dependencies, as a by-product of shorting dependency distance?

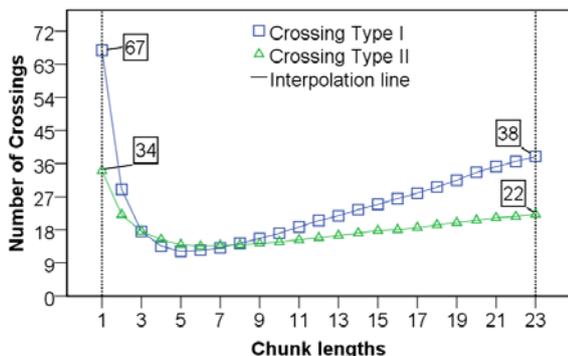

Fig. 6: Relations between the number of crossings and chunk lengths (the length of sentence is 23).

Fig. 6 presents the effects of chunking on the number of dependency crossings. The senence length is 23; chunk size ranges randomly between [1, n](n=1, ..., 23). Reference lines (x=1 and x=23) mark up the number of crossings when n=1 and n=23, which serve as the baseline for comparison. As can be seen in Fig. 6, without any chunking, there are averagely 67 type-I crossings ( similar to the theoretical value obtained by Ferrer-i-Cancho [8-9]) and 34 type-II crossings. When chunks are gernerated with size randomly ranging within [1, 23], there are averagely 38 type-I crossings and 22 type-II crossings for every sentence.

When the maximal chunk length is set as 5, the number of type-I crossings drops to the minimum of 12.19, whereas the number of type-II crossings remains 13.98. With the maximal chunk length set as 6, the number of type-II crossings drops to the minimum of 13.58, whereas the number of type-II crossings remains 12.48. Fig. 6 shows that when sentence length is 23, the minimal MDD can be achieved if chunk size is set somewhere betwee 4 and 7. Therefore, it may be inferred that there is a relation between dependency distance and crossing dependencies, especially the type-I crossings.

Our study reveals that chunking does reduce dependency crossings. Intuitively, the less nodes one chunk has, the more likely it is for intra-chunk crossings to occur——when there are less than 4 nodes in one chunk, intra-chunk type-I crossing is impossible. However, too small chunks will lead to too many chunks, which in turn increase the possibility of inter-chunk type-I crossings. Type-II crossings will not occur if number of noes is less than 3. Similar to type-I crossing, type-II crossing also has much to do with dependency distance. When dependency distance is shorter, dependency is less likely to cross the root. Chunking reduces dependency distance, and thus reduces the likelihood for type-II crossings to appear. Therefore, for sequences with 23 nodes, a chunk length of 6 is probably the optimal balance between the length of chunk and the number of chunks, which can best reduce the number of crossings.

But one thing is evident: the rarity of crossing dependencies is probably by-product of the pressure for short dependency distance, and chunking may play a significant role in DDM. In this study, the computer simulation indicates that, for a sequence with 23 nodes, chunking, though permitting dependency crossings, is still capable of reducing MDD to 5.34, which just a little longer than the MDD (3.84) when dependency crossings are banned. When the chunk length is limited within a certain range, the MDD very often drops below 4, just the same as the MDD found in many natural languages [5]. Existent researches suggest that DDM is probably universal in human languages, as constrained by limited cognitive resources [4-11]. Our study reveals that chunking, which is also universal in human languages, is probably one import mechanism that makes for short MDD in human languages. What is interesting and worthy of further study is probably the relation between chunk and cognition: the size of chunk, which has much to do with MDD, may also bear on human cogntion. However, chunking does not necessarily insure the reduction of dependency crossings to that rarity as found in human languages. In other words, the rarity of dependency crossings, though closely related to short dependency distance, is not an entire by-product of short dependency distance. Behind the rarity of dependency crossings in human languages, there are perhaps more motivations than the mere pressure for short dependency distance. We thereby arrive at the following conclusion: crossing dependencies are costly in languages processing because of not only frequent long distances, but also other inherent properties——even crossings of short dependencies are by and large forbidden in most languages.

What is noteworthy is that our study chunks the sequences at only one stratum, while the sentences in human languages are mostly chunked at different hierarchical levels. Internally, a chunk may be divided into more sub-chunks; externally, a chunk may combine with other chunks to form bigger super-chunks. It seems not unreasonable, therefore, to assume that sub-chunks within chunks may further reduce the number of intra-chunk crossing dependencies, and that super-chunks above chunks may further reduce the number of inter-chunk crossings, and, ultimately, the number of crossings. In view of this possibility, the failure in our study to reduce dependency crossings to such rarity as in human languages may have much to do with the single-stratus chunking we adopted, which does not fully simulate the hierarchical chunking of human languages.

**Conclusion.** – Our study presents the following findings. Firstly, for random sentences whose length is over 20, the constraint of continuity alone cannot reduce their MDD to that of natural langauges. However, for chunked randoms sentences constrained by continuity, their MDD is very close to that of natural languages.

Therefore, chunking may be an important means to realize DDM in a linear sequence whose length is over 20. But mere chunking cannot realize DDM if chunk size is not constrained. This study has oberved that certian chunking sizes are optimal in reducing DD in random sentences. However, it remains a big question whether we could have similar obervations in natural languages.

Secondly, chunking at only one level, which is what we adopt in study, may also significantly reduce the number of dependency crossings. For random sentences whose length is 23, there are averagely 67 type-I crossings and 37 type-II crossings if the sentences are unchunked. However if these sentences are chunked with the size interval [1, 23], the numbers of these two type of crossings are reduced to 38 and 22.

What is interesting is that chunking in our study failed to reduce dependency crossings to such rarity as found in in natural human languages, which very often simply forbid such crossings (practically zero in certain languages [5,7]). This, we assume, may have to do with the one-stratum chunking algorithm adopted in our study.

Thirdly, chunking contributes significantly to the low MDD of human languages. Since low MDD is closely related to the rarity of dependency crossings[4,5,7-10], it may be assumed that the rarity of dependency crossings also has much to do with chunking. As to natural languages, it is possible that multiple factors, including the duality of language, may contribute to the rarity of dependency crossings, expecially.

Chunking is closely related to the defining properties of human languages and the hierarchical structure of a dependency tree, reflecting, perhaps, inherent constraint of our language system on linearization of linguistic symbols. Network relations processed in language network in our brain may translate into hierarchical relations and finally get embodied in terms of chunks in actual linear expression. In this sense, chunking may be one of those fundamental principles in language linearization, that is, it is itself a basic part of syntax. What are close in our mind are often arranged, by chunking, into linear neighborhood, which restrains the linear distance between those related items, on one hand, and reduces the likelihood for dependencies to cross each other, on the other hand, and both of them are to the advantage of language processing.

∗∗∗

This work is supported by the National Social Science Foundation of China (Grant No. 11&ZD188).

REFERENCES

[1] SAUSSURE, F. DE., *Course in General Linguistics*. (Peter Owen, London) 1916/1960.
[2] HUDSON R., *An introduction to word grammar* (Cambridge University Press, Cambridge) 2010.
[3] MEL'ČUK I., *Dependency syntax: theory and practice* (SUNY, Albany) 1988.
[4] LIU H., *Glottometrics*, 15(2007), 1.
[5] LIU H., *J. Cognitive Science*, 9 (2008) 159.
[6] JIANG J. and LIU H., *Lang. Sci.*, 50(2015), 93.
[7] FERRER I CANCHO R., *EPL*, 76(2006) 1228.
[8] FERRER I CANCHO R., *Glottometrics*, 25(2013) 1.
[9] FERRER I CANCHO R., *EPL*, 108 (2014) 58003.
[10] FERRER I CANCHO R. and Gómez-Rodríguez C., arXiv:1508.06451, (2015).
[11] FUTRELL R., MAHOWALD K. and GIBSON E., *PNAS*, 112(2015), 10336.
[12] GILDEA D. and TEMPERLEY D., *Cognitive Science*, 34(2010) 286.
[13] TEMPERLEY D., *Cognition*, 105 (2007) 300.
[14] LECERF Y., *Rapport CETIS No. 4*, (1960) 1 euratom.
[15] HAYS D., *Language*, 40 (1964) 511.
[16] HUDSON R., http://www.phon.ucl.ac.uk/home/dick/difficulty.htm, (1995).
[17] ZIPF G., *Human Behavior and the Principle of Least Effort*. (Addison-Wesley Press, Cambridge, Mass.) 1949.
[18] JACKENDOFF R., *Foundations of language* (Oxford University Press, Oxford) 2002.
[19] MILLER G. A., *Psychol. Rev.*, 63(1956) 81.
[20] Abney S. P., in *Principle-Based Parsing*, edited by BERWICK R. C., Abney S. P. and TENNY C., (Springer Netherlands) 1992, pp. 257-278.
[21] LIU H., *Dependency Grammar: from theory to practice*. (Science Press, Beijing) 2009.
[22] LIU H. and HU F., *EPL*, 83(2008) 18002.
[23] WILSON D. B., In *STOC'96,* ACM, Philadelphia, PA, USA, 1996, pp. 296-303.
[24] QIU L., ZHANG Y., JIN P. and WANG H., in *Proceedings of the 25th COLING*. Dublin, Ireland, 2014, pp. 257-268.
[25] BUK S. and ROVENCHAK A., arXiv:cs/0701194, (2007).